\documentclass[runningheads]{llncs}
\usepackage{graphicx}

\usepackage{tikz}
\usepackage{float}
\usepackage{comment}
\usepackage{amsmath,amssymb} 
\usepackage{color}
\usepackage{bbding}
\usepackage{ulem}

\usepackage[accsupp]{axessibility}  

\begin{document}
\pagestyle{headings}
\mainmatter
\def\ECCVSubNumber{100}  

\title{PanoFormer: Panorama Transformer for Indoor 360° Depth Estimation} 

\titlerunning{PanoFormer}
%
\author{Zhijie Shen\inst{1,2} \and
Chunyu Lin\inst{1,2} \and
Kang Liao\inst{1,2,3}\and
Lang Nie\inst{1,2}\and
Zishuo Zheng\inst{1,2}\and
Yao Zhao\inst{1,2}}
\authorrunning{Shen et al.}
%
\institute{Institute of Information Science, Beijing Jiaotong University, China \and
Beijing Key Laboratory of Advanced Information Science and Network Technology \and
Max Planck Institute for Informatics, Germany\\ 
Corresponding Author: Chunyu Lin\\
\email{\{zhjshen, cylin, kang\_liao, nielang, zszheng, yzhao\}@bjtu.edu.cn}\\
Code: \url{https://github.com/zhijieshen-bjtu/PanoFormer}
}
\maketitle
\begin{abstract}
Existing panoramic depth estimation methods based on convolutional neural networks (CNNs) focus on removing panoramic distortions, failing to perceive panoramic structures efficiently due to the fixed receptive field in CNNs. This paper proposes the \uline{pano}rama trans\uline{former} (named \textbf{\textit{PanoFormer}}) to estimate the depth in panorama images, with tangent patches from spherical domain, learnable token flows, and pano-rama specific metrics. In particular, we divide patches on the spherical tangent domain into tokens to reduce the negative effect of panoramic distortions. Since the geometric structures are essential for depth estimation, a self-attention module is redesigned with an additional learnable token flow. In addition, considering the characteristic of the spherical domain, we present two panorama-specific metrics to comprehensively evaluate the panoramic depth estimation models' performance. 
Extensive experiments demonstrate that our approach significantly outperforms the state-of-the-art (SOTA) methods. Furthermore, the proposed method can be effectively extended to solve semantic panorama segmentation, a similar pixel2pixel task.
\end{abstract}

\section{Introduction}

Depth information is important for computer systems to understand the real 3D world. Monocular depth estimation has attracted researchers'~\cite{Cheng2018Cube,cheng2020omnidirectional,cohen2018spherical,esteves2018learning,yan2021rignet,yan2022multi} attention with its convenience and low cost, especially for panoramic depth estimation~\cite{yu2017flat2sphere,yun2022improving,wang2020bifuse}, where the depth of the whole scene can be obtained from a single 360° image.

 Since estimating depth from a single image is an ill-posed and inherently ambiguous problem, current solutions almost use powerful CNNs to extract explicitly or implicitly prior geometric to realize it~\cite{bhoi2019monocular,chen2021distortion}. However, when applied to panoramic tasks, these SOTA depth estimation solutions for perspective imagery~\cite{laina2016deeper} show a dramatic degradation because the 360° field-of-view (FoV) from panorama brings geometric distortions that challenge the structure perception. Specifically, distortions in panoramas (usually represented in equirectangular projection---ERP) increase from the center to both sides along the latitude direction, severely deforming objects' shapes. Due to the fixed receptive field, CNNs are 
inferior for dealing with distortions and perceiving geometric structures in panoramas~\cite{chen2021distortion}.
\begin{figure}[t]
  \centering
  \includegraphics[width=1\textwidth]{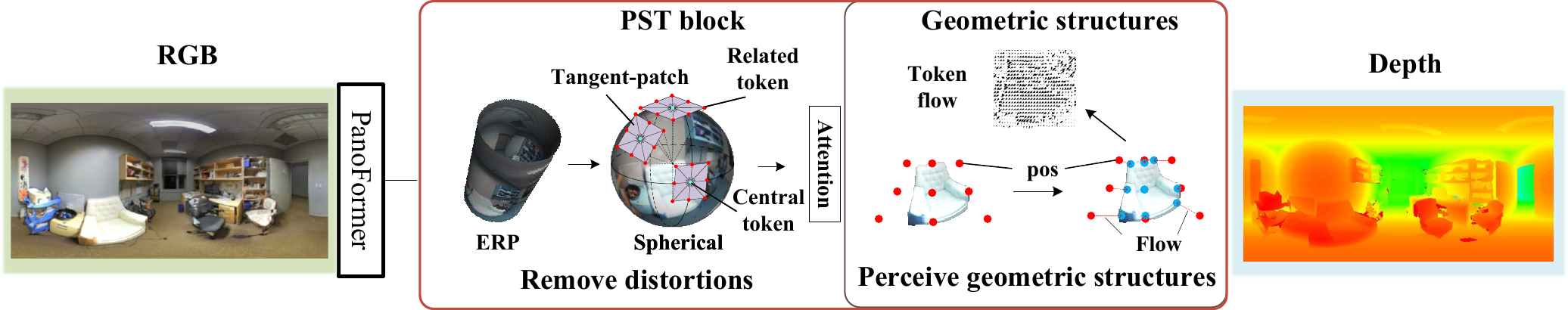} 
  \caption{We present PanoFomer to establish panoramic perception capability. 
  The tangent-patch is proposed to remove panoramic distortions, and the token flows force the token positions to fit the structure of the sofa better.
  More details refer to Sec. \ref{section3}} 
  \label{fig:motivation}
\end{figure}
 To deal with the distortions in panoramas, some researchers~\cite{jiang2021unifuse,shen2021distortion,song2017semantic,wang2020bifuse} adopt the projection-fusion strategy. But this strategy needs to cover the domain gap between different projections, and the extra cross-projection fusion module increases computational burdens. Other researchers~\cite{coors2018spherenet,dai2017deformable,eigen2014depth,su2019kernel,zhu2020deformable,su2019kernel,xiong2018snap,xu2021spherical,jiang2019spherical,khasanova2019geometry,lee2019spherephd} employ various distortion-aware convolution filters to make CNN-based depth estimation solutions adapt to 360° images. However, the fixed sampling positions 
 still limit their performance. Pintore $et\ al.$~\cite{pintore2021slicenet} focuses on the full geometric context of an indoor scene, proposing SliceNet but losing detailed information when reconstructing the depth map. We note that all the existing methods cannot perceive the distorted geometric structures with the fixed receptive field.

To address the above limitations, we propose the first panorama Transformer (PanoFormer) to enable the network's panoramic perception capability by removing distortions and perceiving geometric structures simultaneously (shown in Fig.~\ref{fig:motivation}). To make the Transformer suitable for panoramic dense prediction tasks (e.g., depth estimation and semantic segmentation), we redesign its structure. First, we propose a dense patches dividing method and handcrafted tokens to catch detailed features. Then, we design a relative position embedding method to reduce the negative effect of distortions, which utilizes a central token to locate the eight most relevant tokens to form a tangent patch (it differs from directly dividing patches on the ERP domain in traditional vision Transformers). To achieve this goal, we propose an efficient spherical token locating model (STLM) to guide the `non-distortion' token sampling process on the ERP domain directly by building the Transformations among the three domains (shown in Fig.~\ref{fig:STLM}). Subsequently, we design a Panoramic Structure-guided Transformer (PST) block to replace the traditional block in a hierarchical architecture. Specifically, we redesign the self-attention module with additional learnable weight to push token flow, so as to flexibly capture various objects' structures. This module encourages the PanoFormer to further perceive geometric structures effectively. In this way, we establish our network's perception capability to achieve panoramic depth estimation. Moreover, the proposed PST block can be applied to other learning frameworks as well.
 
Furthermore, current evaluation metrics for depth estimation are suitable for perspective imagery. However, these metrics did not consider distortions and the seamless boundary property in panoramas. To comprehensively evaluate the depth estimation for panoramic images, we design a Pole Root Mean Square Error (P-RMSE) and Left-Right Consistency Error (LRCE) to measure the accuracy on polar regions and depth consistency around the boundaries, respectively.

 Extensive experiments demonstrate that our solution significantly outperforms SOTA algorithms in panoramic depth estimation. Besides, our solution achieves the best performance when applied to semantic segmentation, which is also a pixel2pixel panoramic task. The contributions of this paper are summarized as follows:
\begin{itemize}
     \item We present \textbf{\textit{PanoFormer}}, the first panorama Transformer, to establish the panoramic perception capability by reducing distortions and perceiving geometric structures for the panoramic depth estimation task.
    \item We propose a PST block that divides patches on the spherical tangent domain and reshapes the self-attention module with the learnable token flow. Moreover, the proposed block can be applied in other learning frameworks.
    \item Considering the difference between Panorama and normal images, we design two new panorama-specific metrics to evaluate the panoramic depth estimation.
    \item Experiments demonstrate that our method significantly outperforms the current state-of-the-art approaches on all metrics. The excellent panorama semantic segmentation results also prove the extension ability of our model.
 \end{itemize}

\section{Related Work}

\subsection{Panoramic Depth Estimation}
There are two main fusion methods to reduce distortions while estimating depth on ERP maps. One is the equirectangular-cube fusion method represented by Bifuse~\cite{wang2020bifuse}, and the other is the dual-cube fusion approach described by Shen~\cite{shen2021distortion}. Specifically, Bifuse ~\cite{wang2020bifuse} propose a two-branch method of fusing equirectangular projection and cube projection, which improves the tolerance of the model to distortions. Moreover, UniFuse~\cite{jiang2021unifuse} also uses a dual projection fusion scheme only at the encoding stage to reduce computation cost. Noting that the single-cube projection method produces significant discontinuities at the cube boundary, Shen $et\ al$~\cite{shen2021distortion} proposed a dual-cube approach based on a 45° rotation to reduce distortions. This class of methods can attenuate the negative effect of distortions, but they need to repeatedly change the projection for fusion, increasing the model's complexity.

 To apply depth estimation models of normal images to panoramas, Tateno $et\ al.$~\cite{tateno2018distortion} obtained exciting results by designing distortion-aware convolution filters to expand the perceptual field. 
 Zioulis $et\ al.$~\cite{zioulis2018omnidepth} demonstrated that monocular depth estimation models trained on conventional 2D images produce low-quality results, highlighting the necessity of learning directly on the 360° domain. Jin $et\ al.$~\cite{jin2020geometric} demonstrated the effectiveness of geometric prior for panoramic depth estimation. Chen $et\ al.$~\cite{chen2021distortion} used strip pooling and deformable convolution to design a new encoding structure for accommodating different degrees of distortions. Moreover, Pintore $et\ al.$~\cite{pintore2021slicenet} proposed SliceNet, a network similar to HorizonNet~\cite{sun2019horizonnet}, which uses a bidirectional Long Short-Term Memory (LSTM) to model long-range dependencies. However, the slicing method ignores the latitudinal distortion property and thus cannot accurately predict the depth near the poles. Besides, ~\cite{bhat2021adabins,dosovitskiy2020image} proved that on large-scale datasets, Transformer-based depth estimation for normal images are superior to CNN.

\subsection{Vision Transformer}
 Unlike CNN-based networks, the Transformer has the nature to model long-range dependencies by global self-attention~\cite{ranftl2021vision}. Inspired by ViT~\cite{dosovitskiy2020image}, researchers have designed many efficient networks that have the advantages of both CNNs and Transformers. To enhance local features extraction, convolutional layers are added into muti-head self-attention (CvT~\cite{wu2021cvt}) and feed-forward network (FFN)  (CeiT~\cite{yuan2021incorporating}, LocalViT~\cite{li2021localvit}) is replaced by locally-enhanced  feed-forward network (LeFF) (Uformer~\cite{wang2021uformer}). Besides, CvT~\cite{wu2021cvt} demonstrates that the padding operation in CNNs implicitly encodes position, and CeiT~\cite{yuan2021incorporating} proposes the image-to-tokens embedding method. Inspired by SwinT~\cite{liu2021swin}, Uformer~\cite{wang2021uformer} proposes a shifted windows-based multi-head attention mechanism to improve the efficiency of the model. But all these solutions are developed based on normal FoV images, which cannot be applied to panoramic images directly. Based on these previous works, we further explore suitable Transformer structure for panoramic images and adapt it to the dense prediction task.
\begin{figure}[t]
  \centering
  \includegraphics[width=0.8\textwidth]{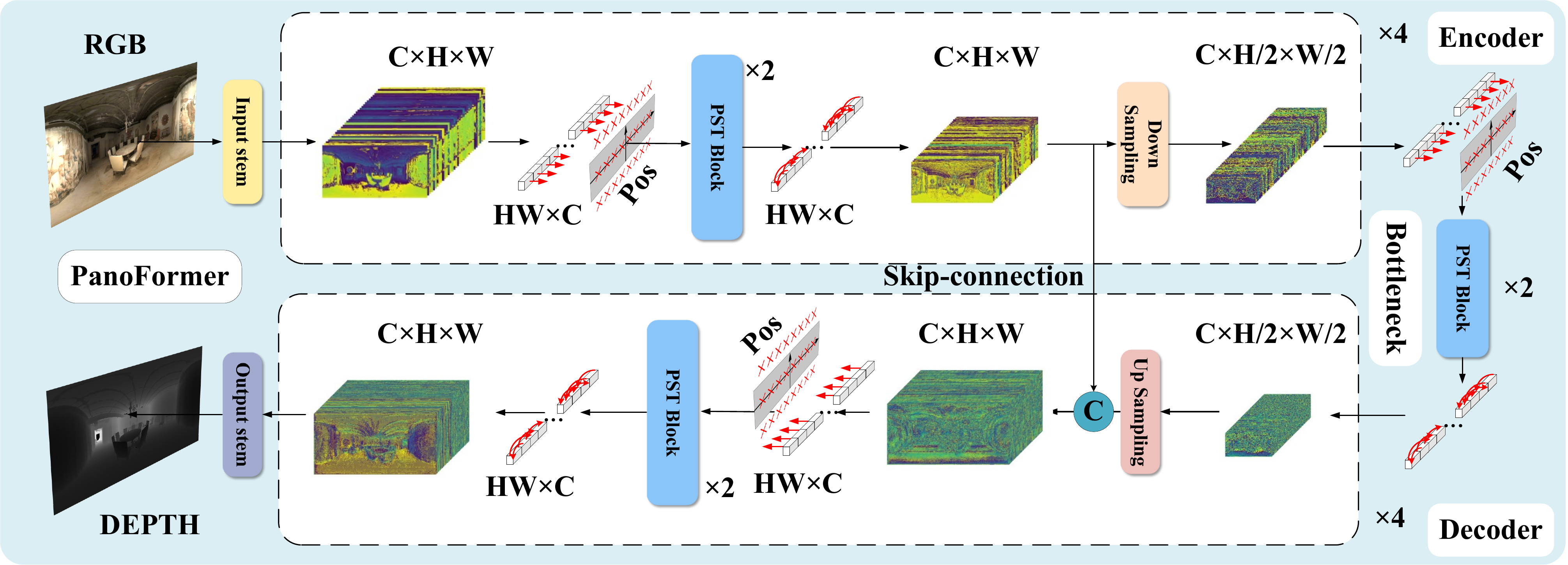} 
  \caption{Our PanoFormer takes a monocular RGB panoramic image as the input and outputs the high-quality depth map} 
  \label{fig:framework}
\end{figure}
\section{PanoFomer}
\label{section3}
\subsection{Architecture Overview}
\label{section30}
 Our primary motivation is to make the Transformer suitable for pixel-level omnidirectional vision tasks by redesigning the standard components in conventional Transformers. Specifically, we propose a pixel-level patch division strategy, a relative position embedding method, and a panoramic self-attention mechanism. The proposed pixel-level patch division strategy is to enhance local features and improve the ability of Transformers to capture detailed features. For position embedding, we renounce the conventional absolute position embedding method and get the position of other related tokens on the same patch by the central token (described in \ref{section33}). This method not only eliminates distortions, but also provides position embedding. Furthermore, we establish a learnable flow in the panorama self-attention module to perceive panoramic structures that are essential for depth estimation. 
 
 As shown in Fig. ~\ref{fig:framework}, the PanoFomer is a hierarchical structure with five major parts: input stem, output stem, encoder, decoder and bottleneck. For the input stem, a 3×3 convolution layer is adopted with size $H$×$W$ to form the features with dimension $C$. Then the features are fed into the encoder. There are four hierarchical stages in encoder and decoder, and each of them contains a position embedding, two PST blocks (sharing the same settings), and a convolution layer. Specifically, a 4×4 convolution layer is adopted for increasing dimension and down-sampling in the encoder, while a 2×2 transposed convolution layer is used in the decoder for decreasing dimension and up-sampling. Finally, the output features from the decoder share the same resolution and dimension as the features from the input stem. Furthermore, the output stem, implemented by a 3×3 convolution, is employed to recover the depth map from features. More specifically, the number of heads is sequentially set as [Encoder:1, 2, 4, 8; Bottleneck: 16; Decoder: 16, 8, 4, 2]. As for all padding operations in convolution layers, we utilize circular padding for both horizontal sides of the features.
  
\subsection{Transformer-customized Spherical Token}
\label{section31}
 In vision Transformers, the input image is first divided into patches of the same size. For example, ViT~\cite{dosovitskiy2020image} divides the input image into patches with size of 16×16 to reduce the computational burden. Then, these patches are embedded as tokens in a learning-based way via a linear layer. However, this strategy loses much detailed information, which is a fatal drawback for dense prediction tasks, such as depth estimation. To overcome this issue, we propose a pixel-level patches dividing method.
 
 First, the input features are divided into pixel-level patches, which means each sampling position in the features corresponds to a patch centered on it. Such a dense division strategy allows the network to learn more detailed features, which is beneficial for dense prediction tasks. Furthermore, we make each patch consist of 9 features at different positions (one central position and eight surrounding positions, illustrated in Fig. \ref{fig:STLM} left) to balance the computational burden. Unlike standard Transformers that embed patches as tokens by a linear layer, our tokens are handcrafted. We define the features at the central position as the central token and those from the other 8 surrounding positions as the related tokens. The central token can determine the position of related tokens by looking up the eight most relevant tokens among the features. To remove distortion and embed position information for the handcrafted tokens, we propose a distortion-based relative position embedding method in Sec. \ref{section32}. 
 
 \subsection{Relative Position Embedding}
 \label{section32}
\begin{figure}[t]
  \centering
  \includegraphics[width=0.8\textwidth]{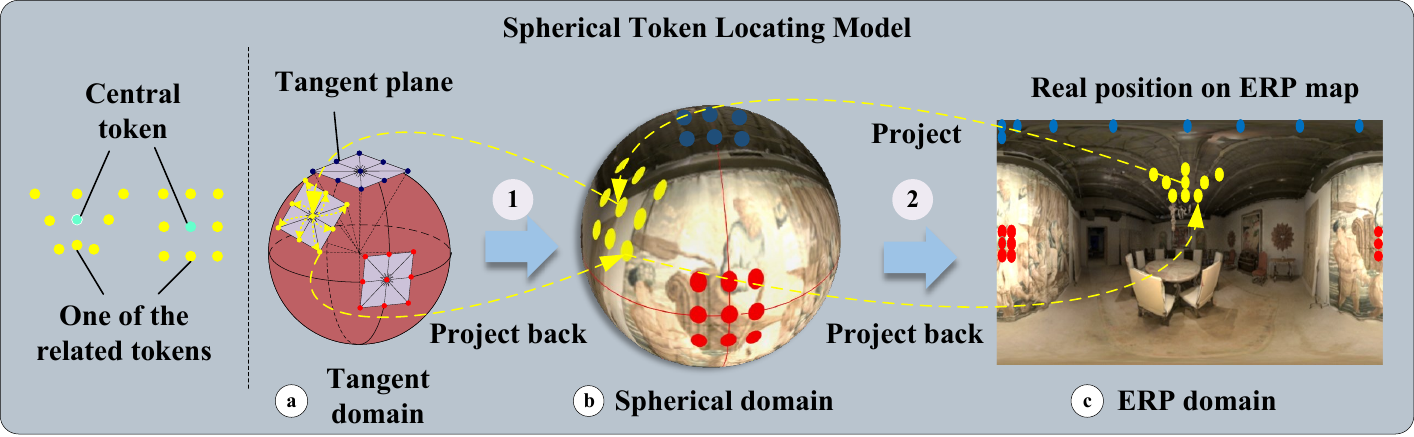} %
  \caption{Spherical Token Locating Model (STLM): locate related tokens on ERP domain. $1$: tangential domain of unit sphere to spherical domain; $2$: spherical domain to ERP domain} %
  \label{fig:STLM}
\end{figure}

 Inspired by the cube projection, we note that the spherical tangent projection can effectively remove the distortion (see Supplementary Materials for proof). Therefore, we propose STLM to initialize the position of related tokens. Unlike the conventional Transformers (e.g., ViT~\cite{dosovitskiy2020image}), which directly adds absolute position encoding to the features, we ``embed" the position information via the central token. Firstly, the central token is projected from the ERP domain to the spherical domain; then, we use the central token to look up the position of eight nearest neighbors on the tangent plane; finally, these positions are all projected back to the ERP domain (the three steps are represented by yellow arrows in Fig. \ref{fig:STLM}). We call patches formed in this way as tangent patches. To facilitate locating the related tokens in the ERP domain, we further establish the relationship among the three domains (illustrated in Fig.~\ref{fig:STLM}).

\noindent \textbf{Tangent domain to spherical domain:} Let the unit sphere be $S^{2}$, and $S(0,0)=(\theta_{0}, \phi_{0} )\in S^{2}$ is the spherical coordinate origin. $\forall S(x,y)=(\theta, \phi )\in S^{2}$, we can obtain other 8 points (related tokens) around it (current token) on the spherical domain.
\begin{eqnarray}
\begin{aligned}
S(\pm 1,0)=&(\theta \pm  \Delta \theta , \phi )\\
S( 0,\pm 1)=&(\theta, \phi \pm  \Delta \phi )\\
S(\pm 1,\pm 1)=&(\theta\pm  \Delta \theta,\phi \pm \Delta \phi  )
\end{aligned}
\end{eqnarray}

where ($\theta, \phi$) denotes the unit spherical coordinates, and $\theta \in (-\pi, \pi)$, $\phi \in (-\frac{\pi }{2}, \frac{\pi }{2})$; $\Delta\theta$,$\Delta\phi$ is the sampling step size.

 By the geocentric projection~\cite{pearson1990map}, we can calculate the local coordinates ($T(x, y)$)of the sampling point in tangent domain~\cite{coors2018spherenet} (the current token in tangent domain is represented as $T(0, 0) = T(\theta, \phi) = (0, 0)$):
 \begin{eqnarray}
\begin{aligned}
T(\theta \pm  \Delta \theta , \phi)=&( \pm \tan \Delta \phi, 0 )\\
T(\theta, \phi \pm  \Delta \phi )=&(0, \phi \pm \tan \Delta \theta )\\
T(\theta\pm  \Delta \theta,\phi \pm \Delta \phi)=&(\pm \tan \Delta \phi, \pm \sec \Delta \phi \tan \Delta \theta  )
\end{aligned}
\end{eqnarray}
By applying the inverse projection described in~\cite{coors2018spherenet}, we can get the position of all tokens of a tangent patch in the spherical domain.

\noindent \textbf{Spherical domain to ERP domain:} Furthermore, by utilizing the projection equation~\cite{shen2021distortion}, we can get the position of each tangent patch in the ERP domain. This whole process is named Spherical Token Locating Model (STLM).

\begin{figure}[t]
  \centering
  \includegraphics[width=0.8\textwidth]{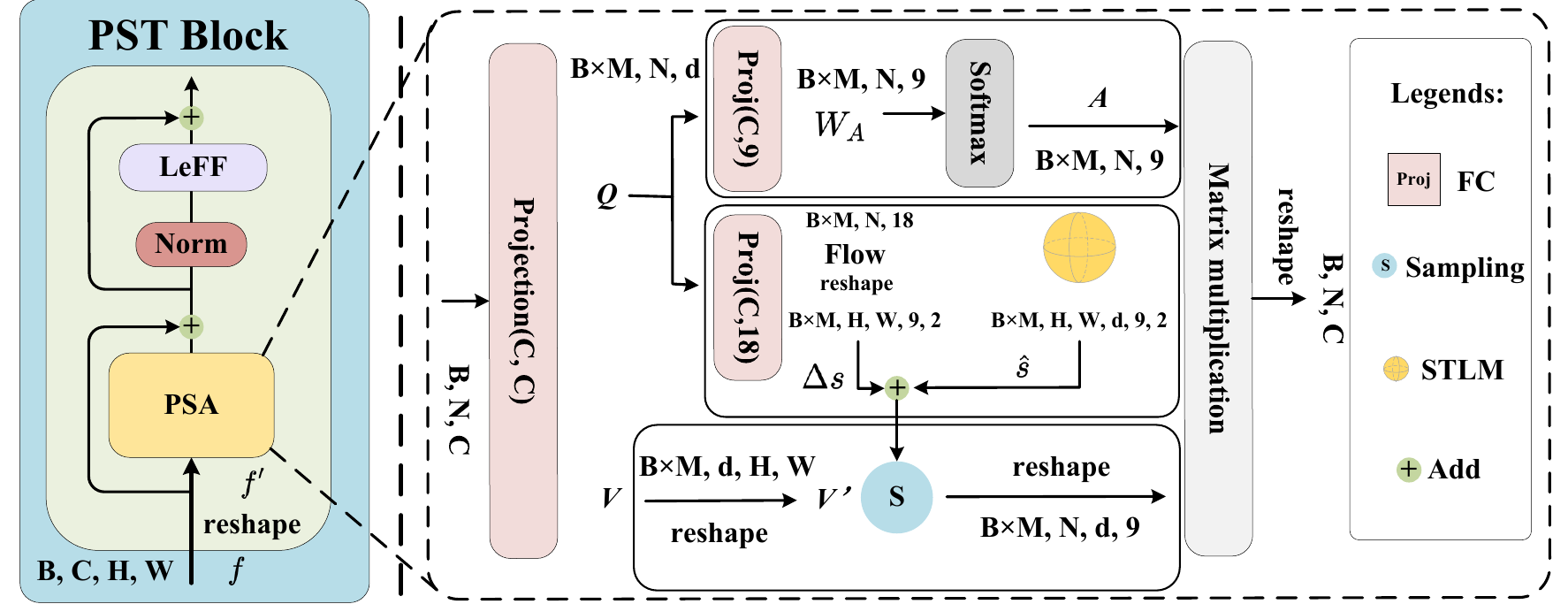} 
  \caption{The proposed PST Block can remove the negative effect of distortions and perceive geometric structures} 
  \label{fig:block}
\end{figure}
\subsection{Panorama Self-Attention with Token Flow}
\label{section33}
Based on the traditional vision Transformer block, we replace the original attention mechanism with panorama self-attention. To further enhance local features interaction, we replace FFN with LeFF~\cite{yuan2021incorporating} for our pixel-level depth estimation task.
Specifically, as illustrated in Fig. \ref{fig:block}, when the features $f \in \mathbb{R}^{C \times H \times W}$ with a height of $H$ and a width of $W$ are fed into PST block, they are flattened and reshaped as $f' \in \mathbb{R}^{N \times C}$, where $N = H \times W$. Then a  fully connected layer is applied to obtain query $Q \in\mathbb{R}^{N \times d}$ and value $V \in\mathbb{R}^{N \times d}$, where $d = C/M$, and $M$ is the head number. The $Q$ and $V$ will pass through three parallel branches for computing attention score ($A\in \mathbb{R}^{N \times 9}$), token flows ($\Delta s \in \mathbb{R}^{N \times 18}$), and re-sampling features. In the top branch, a full connection layer is adopted to get attention weights $W_{A} \in \mathbb{R}^{N \times 9}$ from $Q$, and then softmax is employed to calculate the attention score $A$. In the middle branch, another fully connection layer is used to learn a token flow $\Delta s$ and it is further reshaped to $\Delta s' \in \mathbb{R}^{d \times H \times W \times 9 \times 2}$, sharing the same dimension with $\hat{s}$ (the initialed position from the STLM). Moreover, $\Delta s'$ and $\hat{s}$ are added together to calculate the final token positions. In the bottom branch, the value $V$ is reshaped to $V' \in \mathbb{R}^{C \times H \times W}$ and are sampled to form the divided patches (described in \ref{section30}) by looking up the related tokens in the final token positions. Afterward, the PSA can be represented as follows:
\begin{eqnarray}
\begin{array}{cc}
&\operatorname{PSA}(f, \hat{\boldsymbol{s}})=
\sum_{m=1}^{M}W_{m}*\\&\left[ \sum_{q=1}^{H\times W}\sum_{k=1}^{9}  A_{m q k} \cdot W_{m}^{\prime} f\left(\hat{s}_{mqk}+\Delta s_{m q k}\right)\right],
\end{array}
\end{eqnarray}
where $\hat{s} = \boldsymbol{STLM}(f)$, and $\boldsymbol{STLM}$($\cdot$) denotes the spherical token locating model; m indexes the head of self-attention, M is the whole heads, q index the current point (token), k indexes the tokens in a tangent patch, $\Delta \boldsymbol{s}_{m q k}$ is the learned flow of each token, 
$A_{m q k}$ represents the attention weight of each token, and $W_{m}$ and $W_{m}^{\prime}$ are normal learnable weights of each head.

\begin{figure}[t]
  \centering
  \includegraphics[width=0.5\textwidth]{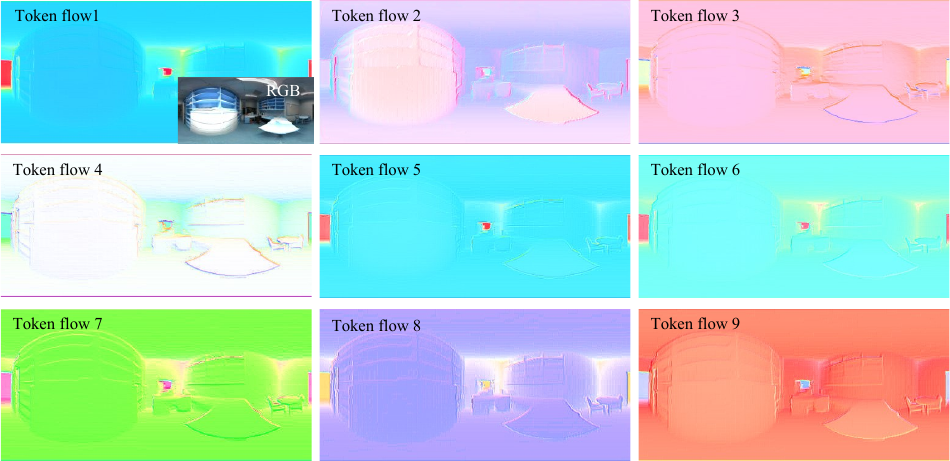} 
  \caption{Visualization of the token flows from the first PST block, which suggest the panoramic structures} 
  \label{fig:pos}
\end{figure}

From the above process, we can see that the final positions of the tokens are determined by two steps: position initialization from STLM and additional learnable flow. Actually, the initialized position realizes the division of the tangent patch (described in \ref{section33}) and removes the panoramic distortion. Furthermore, the learnable flow exhibits a panoramic geometry by adjusting the spatial distribution of tokens. To verify the effectiveness of the token flow, we visualize all tokens from the first PST block in Fig. \ref{fig:pos}. It can be observed that this additional flow provides the network with clear scene geometric information, which helps the network to estimate the panorama depth with the structure as a clue.
 
\subsection{Objective Function}
\label{section34}
For better supervision, we combine reverse Huber~\cite{esmaeili2019novel} (or Berhu~\cite{laina2016deeper}) loss and gradient loss~\cite{shen2021distortion} to design our objective function as commonly used in previous works~\cite{pintore2021slicenet,shen2021distortion}. In our objective function, the Berhu loss $\beta_{\delta}$ can be written as:
\begin{eqnarray}
\beta_{\delta}(g, p)=\left\{\begin{array}{ll}
|g-p| & \text { for }|g-p| \leq \delta \\
\frac{|(g-p)^{2}|+ \delta^{2}}{2\delta} & \text { otherwise }
\end{array}\right.
\end{eqnarray}
where $g, p$ denote the ground truth and predicted values, respectively.

Similar to SliceNet~\cite{pintore2021slicenet}, we apply gradient loss to Berhu loss. To obtain depth edges, we use two convolution kernels to obtain gradients in horizontal and vertical directions, respectively. They are represented as $K_{h}$ and $K_{v}$, where $K_{h}$ = [-1 0 1, -2 0 2, -1 0 1], and $K_{v} = (K_{h})^T$. Denote the gradient function as $G$, the horizontal gradient $I_{h}$ and vertical gradient $I_{v}$ of the input image $I$ can be expressed as $I_{h} = G(K_{h}, I)$ and $I_{v} = G(K_{v}, I)$, respectively.
In this paper, $\delta$ = 0.2 and the final objective function can be written as
\begin{eqnarray}
\ell_{final} = \beta_{0.2} (g,p) + \beta_{0.2} (G(K_{h}, g), G(K_{h}, p)) + \beta_{0.2} (G(K_{v}, g), G(K_{v}, p)),
\end{eqnarray}
\section{Panorama-specific Metrics}
\label{section4}
Rethinking the spherical domain, we note that two significant properties cannot be neglected: the spherical domain is continuous and seamless everywhere; the distortion in the spherical domain is equal everywhere. For the first issue, we propose LRCE to measure the depth consistency of left-right boundaries. For the second issue, since distortions on ERP maps vary in longitude, RMSE cannot visually reflect the model's ability to adapt to distortions. Therefore, we provide P-RMSE to focus on the regions with massive distortions to verify the model's panoramic perception capability.
\begin{figure}[H]
  \centering
  \includegraphics[width=0.4\textwidth]{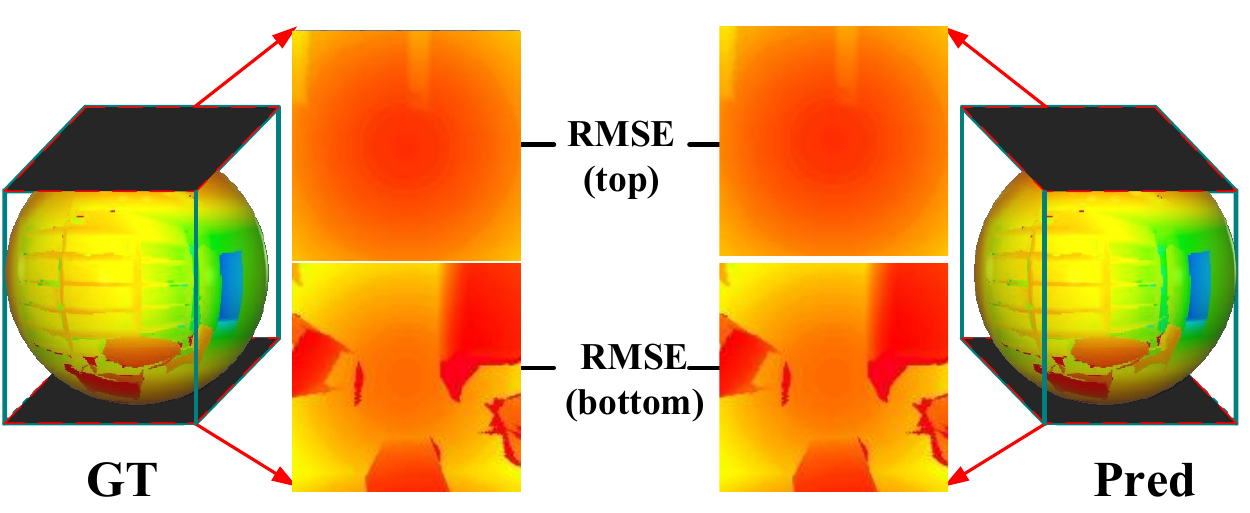} 
  \caption{P--RMSE: calculate the RMSE of the polar regions} 
  \label{fig:prmse}
\end{figure}
\noindent \textbf{Pole Root Mean Square Error.} Cube projection is a special spherical tangent projection format that projects the sphere onto the cube's six faces. The top and bottom faces correspond to the polar regions of the spherical domain, so we select the two parts to design P-RMSE (illustrated in Fig. ~\ref{fig:prmse}). Define the function of converting ERP to Cube as $E2C(\cdot)$, the converted polar regions of the ERP image $E$ can be expressed as $Select(E2C(E), T, B)$, where $T, B$ represent the top and bottom parts, respectively. The error $C_{e}$ between the ground truth $GT$ and the predicted depth map $P$ at the polar regions can be expressed as  
\begin{eqnarray}
C_{e}&=&Select(E2C(GT), T, B) - Select(E2C(P), T, B)
\end{eqnarray}
The final P-RMSE can be written as
\begin{eqnarray}
\operatorname{P-RMSE}&=&\sqrt{\frac{1}{N_{C_{e}}}\sum_{i=1}^{N_{C_{e}} }|C_{e}^{i}|  } 
\end{eqnarray}
where $N_{C_{e}}$ is the number of values in $C_{e}$.

\noindent \textbf{Left-Right Consistency Error.}
We can evaluate the depth consistency of the left-right boundaries by calculating the horizontal gradient between the both sides of the depth map. Define that the horizontal gradient $G_{E}^{H}$ of the image $E$ can be written as $G_{E}^{H} = E^{col}_{first} - E^{col}_{last}$, where $E^{col}_{first}$/ $E^{col}_{last}$ represent the values in the first/last columns of the image $E$. But consider an extreme case where if the edge of an object in the scene happens to be on the edge of the depth map, then there is ambiguity in reflecting continuity only by $G_{E}^{H}$. We cannot tell whether this discontinuity is real or caused by the model. Therefore, we add ground truth to our design. The horizontal gradient of ground truth and the predicted depth map are denoted as $G_{GT}^{H}$ and $ G_{P}^{H}$ (where $G_{GT}^{H} = GT^{col}_{first} - GT^{col}_{last}$, $G_{P}^{H} = P^{col}_{first} - P^{col}_{last}$), respectively. The final expression can be as follows:
\begin{eqnarray}
\operatorname{LRCE} = \frac{1}{N_{error} }\sum_{i=1}^{N_{error}}|error_{i}|
\end{eqnarray}
where $error = G_{GT}^{H} - G_{P}^{H}$ and $N_{error}$ is the number of values in $error$.
\begin{table*}[t]
\begin{center}
\caption{Quantitative comparisons on Matterport3D, Stanford2D3D, PanoSUNCG and 3D60 Datasets.} \label{tab:cp1}
\resizebox{\textwidth}{!}{\begin{tabular}{c|c|c|c|c|c|c|c}
\hline
  &&\multicolumn{6}{c}{Classic metrics}\\
 \cline{3-8}
  Dataset & Method &$\delta_{1}\uparrow$&$\delta_{2}\uparrow$&$\delta_{3}\uparrow$& \textbf{RMSE}$\downarrow$&MRE$\downarrow$&MAE$\downarrow$
  \\
  \cline{3-8}
  &&\multicolumn{3}{c|}{Higher the better}&\multicolumn{3}{c}{Lower the better}\\
  \hline
   & FCRN~\cite{laina2016deeper}&0.7703&0.9174&0.9617&0.6704&0.2409&0.4008\\
  \cline{2-8}
  & OmniDepth~\cite{zioulis2018omnidepth}&0.6830&0.8794&0.9429&0.7643&0.2901&0.4838 \\
  \cline{2-8}
    Matterport3D&Bifuse~\cite{wang2020bifuse}&0.8452&0.9319&0.9632&0.6295&0.2408&0.3470\\
  \cline{2-8}
  & UniFuse~\cite{jiang2021unifuse}&0.8897&0.9623&0.9831&0.4941&--&0.2814 \\
   \cline{2-8}
   & SliceNet~\cite{pintore2021slicenet}&0.8716&0.9483&0.9716&--&0.1764&0.3296 \\
   \cline{2-8}
    & Ours &\textbf{0.9184}&\textbf{0.9804}&\textbf{0.9916}&\textbf{0.3635}&\textbf{0.0571}&\textbf{0.1013}\\
 \hline
   & FCRN~\cite{laina2016deeper}&0.7230&0.9207&0.9731&0.5774&0.1837&0.3428\\
    \cline{2-8}
 & OmniDepth~\cite{zioulis2018omnidepth} &0.6877&0.8891&0.9578&0.6152&0.1996&0.3743 \\
  \cline{2-8}
   Stanford2D3D&Bifuse~\cite{wang2020bifuse}&0.8660&0.9580&0.9860&0.4142&0.1209&0.2343\\
  \cline{2-8}
  & UniFuse~\cite{jiang2021unifuse}&0.8711&0.9664&0.9882&0.3691&--&0.2082 \\
   \cline{2-8}
   & SliceNet~\cite{pintore2021slicenet}&0.9031&0.9723&0.9894&--&0.0744&0.1048 \\
   \cline{2-8}
    &Ours &\textbf{0.9394}&\textbf{0.9838}&\textbf{0.9941}&\textbf{0.3083}&\textbf{0.0405}&\textbf{0.0619}\\
  \hline
  \hline
   &&\multicolumn{4}{c|}{Classic metrics}&\multicolumn{2}{c}{New metrics}\\
 \cline{3-8}
  Dataset & Method &$\delta_{1}\uparrow$&$\delta_{2}\uparrow$&$\delta_{3}\uparrow$& \textbf{RMSE}$\downarrow$&P-RMSE$\downarrow$&LRCE$\downarrow$
  \\
  \hline
    & FCRN~\cite{laina2016deeper}&0.9532&0.9905&0.9966&0.2833&0.1094&0.1119\\
    \cline{2-8}
 & OmniDepth~\cite{zioulis2018omnidepth} &0.9092&0.9702&0.9851&0.3171&0.0929&0.0913 \\
  \cline{2-8}
   PanoSUNCG&Bifuse~\cite{wang2020bifuse}&0.9590&0.9838&0.9907&0.2596&0.0967&0.0735\\
  \cline{2-8}
  & UniFuse~\cite{jiang2021unifuse}&0.9655&0.9846&0.9912&0.2802&0.0826&0.0884 \\
   \cline{2-8}
    &Ours &\textbf{0.9780}&\textbf{0.9961}&\textbf{0.9987}&\textbf{0.1503}&\textbf{0.0537}&\textbf{0.0442}\\
  \hline
    & FCRN~\cite{laina2016deeper}&0.9532&0.9905&0.9966&0.2833&0.1681&0.2100\\
    \cline{2-8}
 & OmniDepth~\cite{zioulis2018omnidepth} &0.9092&0.9702&0.9851&0.3171&0.1373&0.1941 \\
  \cline{2-8}
   &Bifuse~\cite{wang2020bifuse}&0.9699&0.9927&0.9969&0.2440&0.1229&0.1357\\
  \cline{2-8}
  3D60& UniFuse~\cite{jiang2021unifuse}&0.9835&0.9965&0.9987&0.1968&0.0829&0.1021 \\
  \cline{2-8}
   & DAMO~\cite{chen2021distortion}&0.9865&0.9966&0.9987&0.1769&--&-- \\
   \cline{2-8}
   & SliceNet~\cite{pintore2021slicenet}&0.9788&0.9952&0.9969&--&0.1746&0.1600 \\
   \cline{2-8}
    & Ours &\textbf{0.9876}&\textbf{0.9975}&\textbf{0.9991}&\textbf{0.1492}&\textbf{0.0501}&\textbf{0.0898}\\
\hline
\end{tabular}}
\end{center}
\end{table*}

\section{Experiments}
\label{section5}
In the experimental part, we compare the state-of-the-art approaches on four popular datasets and validate the effectiveness of our model.
\subsection{Datasets and Implementations}
Four datasets are used for our experimental validation, they are Stanford2D-3D~\cite{armeni2017joint}, Matterport3D~\cite{chang2017matterport3d}, PanoSUNCG~\cite{song2017semantic} and 3D60~\cite{zioulis2018omnidepth}. 

Stanford2D3D and Matterport3D are two real-world datasets. They were rendered from a common viewpoint. Previous work used a dataset that was rendered only on the equator and its surroundings, ignoring the area near the poles, which undermined the integrity of the panorama. 
We strictly follow the previous works and employ the official datasets (Notice that the Stanford2D3D and Matterport3D that are contained in 3D60 have a problem that the light in the scenarios will leak the depth information).
PanoSUNCG is a virtual panoramic dataset. And 3D60 is an updated version of 360D (360D is no longer available now). It consists of data from the above three datasets. There is a gap between the distributions of these three datasets, which makes the dataset more responsive to the model's generalizability. Note that we divide the dataset as the previous work and eliminate the samples that failed to render~\cite{chen2021distortion,wang2020bifuse}.

In the implementation, we conduct our experiments on two GTX 3090 GPUs, and the batch size is set to 4. We choose Adam~\cite{kingma2014adam} as the optimizer and keep the default settings. The initialized learning rate is $1\times10^{-4}$. The number of parameters of our model is 20.37 M.
\begin{figure}[t]
  \centering
  \includegraphics[width=0.8\textwidth]{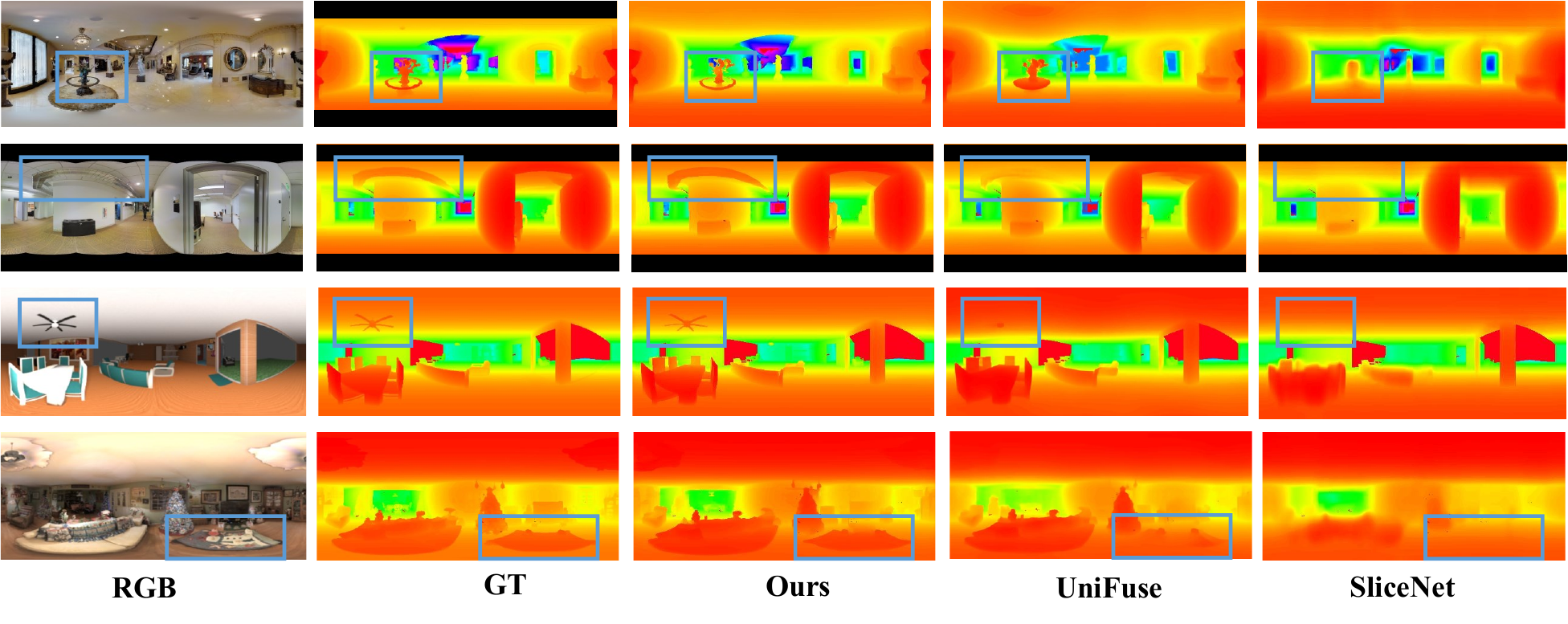} 
  \caption{Qualitative results on Matterport3D, Stanford2D3D, PanoSUNCG, and 3D60. More results can be found in Supplementary Materials} 
  \label{fig:cp2}
\end{figure}
\begin{figure}[!t]
  \centering
  \includegraphics[width=0.8\textwidth]{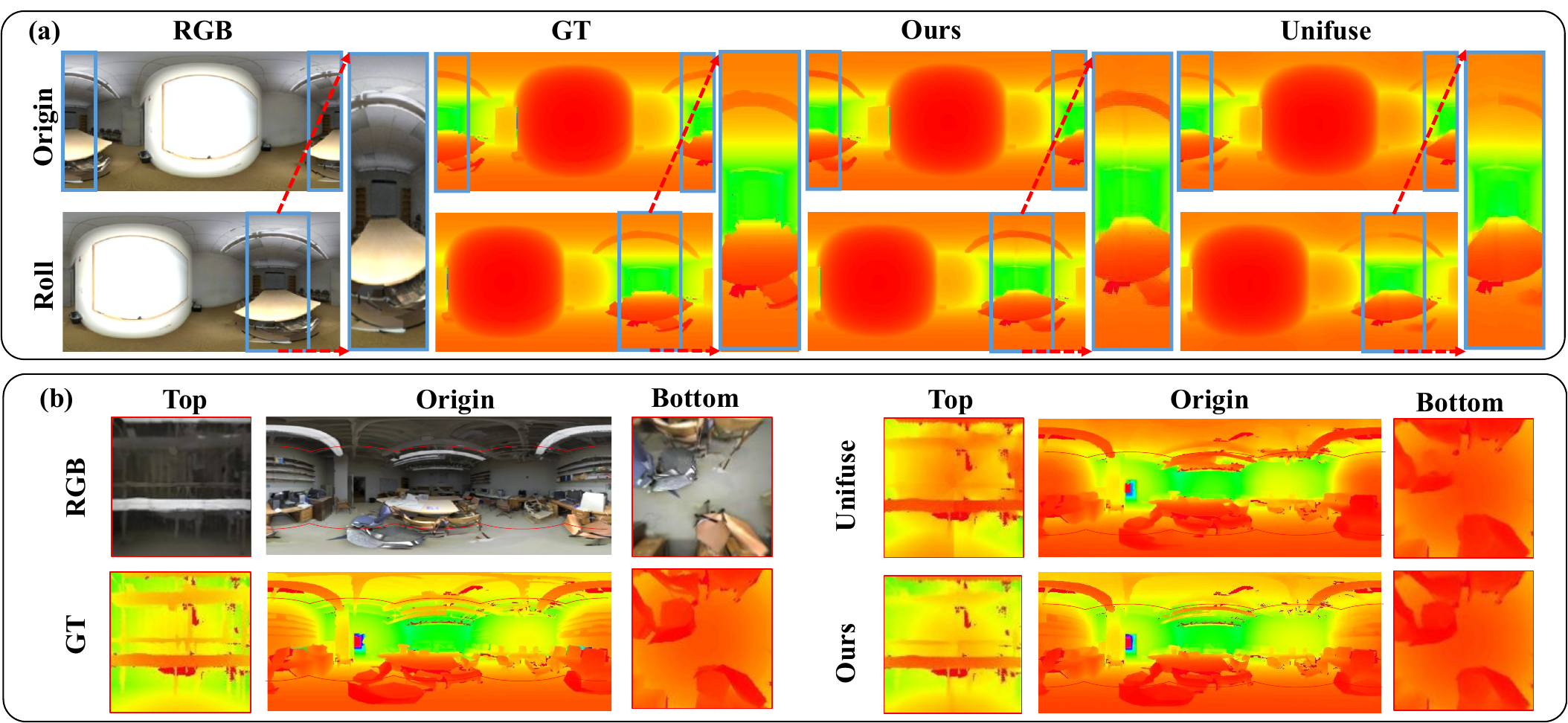} 
  \caption{Visualization of the new metrics' comparison between our method and Unifuse~\cite{jiang2021unifuse}. (a) We stitch the ERP results to observe the depth consistency. (b) We project the areas with massive distortions to cube face to compare the models' performance} 
  \label{fig:metrics}
\end{figure}
\subsection{Comparison Results}
We selected the metrics used in previous work and the two proposed metrics for the quantitative comparison, including RMSE, $\delta$($1.25$, $1.25^{2}$, $1.25^{3}$) and panorama-specific metrics, LRCE and P-RMSE (We cannot calculate the proposed new metrics due to limitation of the two real-world datasets). RMSE reflects the overall variability. $\delta$ exhibits the difference between ground truth and the predicted depth. 

\noindent \textbf{Quantitative Analysis.}
Table ~\ref{tab:cp1} shows the quantitative comparison results with the current SOTA monocular panoramic depth estimation solutions on the four popular datasets. 
As shown in the table, our model achieves the first place in all metrics. In particular, the RMSE metric of our model achieves a 16\% improvement on Stanford2D3D, 26\% on Matterport3D. Even on the virtual dataset PanoSUNCG, there is a 42\% improvement on RMSE. But there is just a 16\% improvement on 3D60 dataset with RMSE. The improvement is not particularly significant compared to the other three datasets because 3D60 dataset is more extensive, the difference between the models is not obvious. The improvement on $\delta$ performance further demonstrates that our model can obtain more accurate prediction results. On the new metric P-RMSE, we achieved an average gain of about 40\% on the other two virtual datasets. It indicates that our model is more resilient to the distortion in panoramas. In addition, on LRCE, our model outperforms 40\% on PanoSUNCG and 12\% on 3D60, showing that our model can better constrain the depth consistency of the left-right boundaries in panoramas, because our network fully considers the seamless property of the sphere.

\noindent \textbf{Qualitative Analysis.} Fig.~\ref{fig:cp2} shows the qualitative comparison with the current SOTA approaches. From the figures, we can observe that SliceNet is relatively accurate in predicting regions without distortion. However, the model performance degrades dramatically in regions with distortions or large object deformations. Although SliceNet can efficiently focus on the global panoramic structures, the depth reconstruction process cannot accurately recover the details, which affects the model's performance.
UniFuse can deal with deformation effectively, but it still suffers from incorrect estimating and tends to lose detailed information. From Fig. ~\ref{fig:metrics}, we can observe that our results are very competitive at boundary and pole areas.

\setlength{\tabcolsep}{5pt}
\begin{table}[!ht]
\begin{center}
\caption{Ablation study. We trained on Stanford2D3D for 70 epochs. $a$ is the baseline structure developed with CNNs}
\label{ablation}
 \begin{tabular}{c|c|c|c|c|c|c}
  \hline
  Index& Transformer &  STLM& Token Flow& RMSE& P-RMSE& LRCE \\
  \hline
  $a$&\XSolidBrush&\XSolidBrush& \XSolidBrush&0.6704&0.2258&0.2733\\
   \hline
   $b$&\Checkmark&\XSolidBrush&\XSolidBrush&0.4349&0.2068&0.2155 \\
  \hline
    $c$&\Checkmark&\Checkmark&\XSolidBrush&0.3739&0.1825&0.1916\\
  \hline
    $d$&\Checkmark&\Checkmark&\Checkmark&\textbf{0.3366}&\textbf{0.1793}&\textbf{0.1784}\\
  \hline
 \end{tabular}
 \end{center}
 \vspace{-2em}
\end{table}
\subsection{Ablation study}
With the same conditions, we validated the key components of our model by ablation study on Stanford2D3D (real-world dataset, small-scale, challenging). As illustrated in Table~\ref{ablation}, $a$ presents the baseline structure that we use convolutional layers to replace PST blocks; Our network with the traditional attention mechanism is expressed with $b$; $c$ indicates our attention module without token flow; Our entire network is shown as $d$.

\noindent \textbf{Transformer vs. CNN.}
From Table \ref{ablation}, we can observe that the Transformer gains 35\% improvements over CNNs in terms of RMSE. Furthermore, qualitative results in $b$ are more precise than the CNNs. Essentially, CNNs are a special kind of self-attention. Since the convolutional kernel is fixed, it requires various components or structures or even deeper networks to help the model learn the data patterns. On the other hand, the attention in Transformer is more flexible, and it is relatively easier to learn the patterns.

\noindent \textbf{Effectiveness of Tangent-patches for Transformer.} 
To illustrate the effectiveness of the tangent-patch dividing method, we compared an alternative attention structure that currently performs SOTA in vision Transformers. From Table ~\ref{ablation}, our network with tangent-patches ($c$) outperforms the attention mechanism ($b$) with 21\% on RMSE, 10\% on P-RMSE and 12\% on LRCE. It proves that tangent-patch can help networks deal with panoramic distortions. 

\begin{figure}[t]
  \centering
  \includegraphics[width=0.8\textwidth]{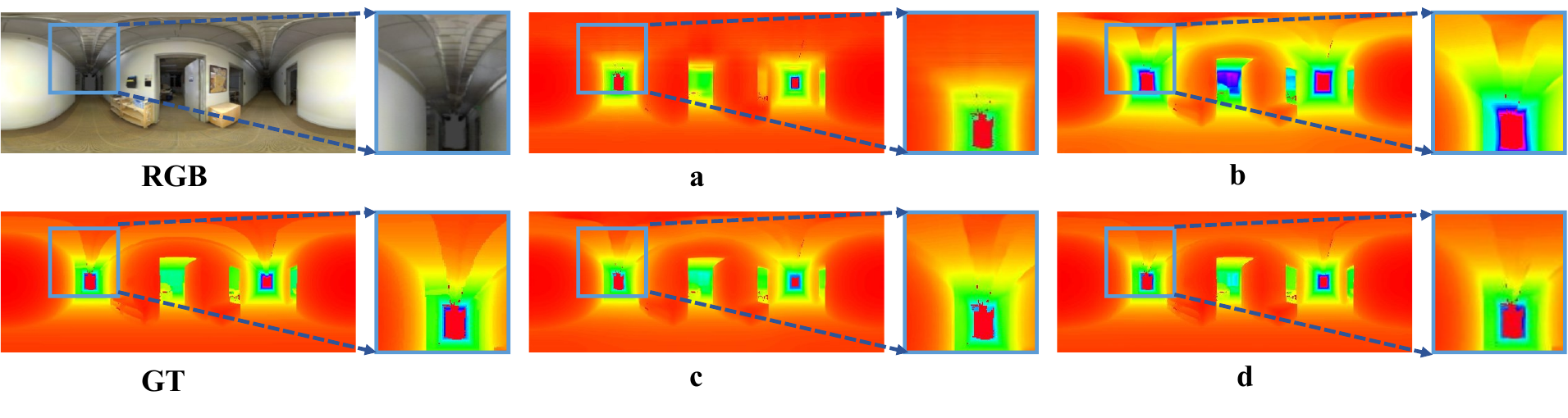} 
  \caption{Qualitative comparison of ablation study. $a, b, c, d$ are the same as Table~\ref{ablation}} 
  \label{fig:ab}
\end{figure}

\noindent \textbf{Effectiveness of Token Flow.} Since the geometric structures are essential for depth estimation, we add the additional token flows to perceive geometric structures. The results in Table~\ref{ablation} show that our model with the token flow can make P-RMSE more competitive. In Fig.~\ref{fig:ab}, we can observe that the token flow allows the model to estimate the depth details more accurately.
\subsection{Extensibility}
 We also validate the extensibility of our model by the panoramic segmentation that is also a pixel2pixel task. We did not change any structure of our network and strictly followed the experimental protocol in ~\cite{sun2021hohonet}. As listed in Table \ref{tab:seg}, the experimental results show that our model outperforms the current SOTA approaches. Due to page limitations, more qualitative comparisons and the results with a high resolution can be found in the supplementary material.
 \begin{table}[ht]
\begin{center}
\caption{Quantitative comparison for semantic segmentation on Stanford2D3D. Results are averaged over the official 3 folds~\cite{sun2021hohonet}}
\label{tab:seg}
\begin{tabular}{c|c|c|c}
  \hline
  Dataset & Method &mIoU$\uparrow$&mAcc$\uparrow$
  \\
  \hline
   & TangentImg~\cite{eder2020tangent}&41.8&54.9\\
   \cline{2-4}
  Stanford2D3D& HoHoNet~\cite{sun2021hohonet}&43.3&53.9\\
   \cline{2-4}
    &Ours&\textbf{48.9}&\textbf{64.5}\\
  \hline
\end{tabular}
\end{center}
\vspace{-2.5em}
\end{table}
\section{Conclusion}
\label{section6}
In this paper, we propose the first panorama Transformer (PanoFormer) for indoor panoramic depth estimation. Unlike current approaches, we remove the negative effect of distortions and further model geometric structures by using learnable token flow to establish the network's panoramic perceptions. Concretely, we design a PST block, which can be effectively extended to other learning frameworks. To comprehensively measure the performance of the panoramic depth estimation models, we propose two panorama-specific metrics based on the priors of equirectangular images. Experiments demonstrate that our algorithm significantly outperforms current SOTA methods on depth estimation and other pixel2pixel panoramic tasks, such as semantic segmentation.

\paragraph{\textbf{Acknowledgement}. This work was supported by the National Key R$\&$D Program of China (No.2021ZD0112100), and the National Natural Science Foundation of China (Nos. 62172032, U1936212, 62120106009).}
\bibliographystyle{splncs04}
\bibliography{egbib}
\end{document}